\documentclass[10pt,twocolumn,letterpaper]{article}

\usepackage{cvpr}              

\usepackage{graphicx}
\usepackage{amsmath}
\usepackage{amssymb}
\usepackage{booktabs}
\usepackage{float}
\usepackage{makecell}
\usepackage{mathtools}
\usepackage{ragged2e}
\usepackage{pifont}
\newcommand{\xmark}{\ding{55}}%
\usepackage{algorithm}
\usepackage{algorithmic}
\usepackage{comment}
\usepackage{enumerate}
\usepackage{multirow}

\usepackage[pagebackref,breaklinks,colorlinks]{hyperref}

\usepackage[capitalize]{cleveref}
\crefname{section}{Sec.}{Secs.}
\Crefname{section}{Section}{Sections}
\Crefname{table}{Table}{Tables}
\crefname{table}{Tab.}{Tabs.}

\def\cvprPaperID{8991} 
\def\confName{CVPR}
\def\confYear{2022}

\begin{document}

\title{HPS-Det: Dynamic Sample Assignment with Hyper-Parameter Search for Object Detection}

\author{Ji Liu, Dong Li, Zekun Li, Han Liu, Wenjing Ke, Lu Tian, Yi Shan \\
 Advanced Micro Devices, Inc., Beijing, China \\
{\tt\small \{ji.liu, d.li, han.liu, wenjing.ke, lu.tian, yi.shan\}@amd.com
}}

\maketitle

\begin{abstract}
Sample assignment plays a prominent part in modern object detection approaches. However, most existing methods rely on manual design to assign positive / negative samples, which do not explicitly establish the relationships between sample assignment and object detection performance. In this work, we propose a novel dynamic sample assignment scheme based on hyper-parameter search. We first define the number of positive samples assigned to each ground truth as the hyper-parameters and employ a surrogate optimization algorithm to derive the optimal choices. Then, we design a dynamic sample assignment procedure to dynamically select the optimal number of positives at each training iteration. Experiments demonstrate that the resulting HPS-Det brings improved performance over different object detection baselines. Moreover, We analyze the hyper-parameter reusability when transferring between different datasets and between different backbones for object detection, which exhibits the superiority and versatility of our method.

\end{abstract}

\begin{figure}[t!]
\footnotesize
\begin{center}
   \includegraphics[width=1.0\linewidth]{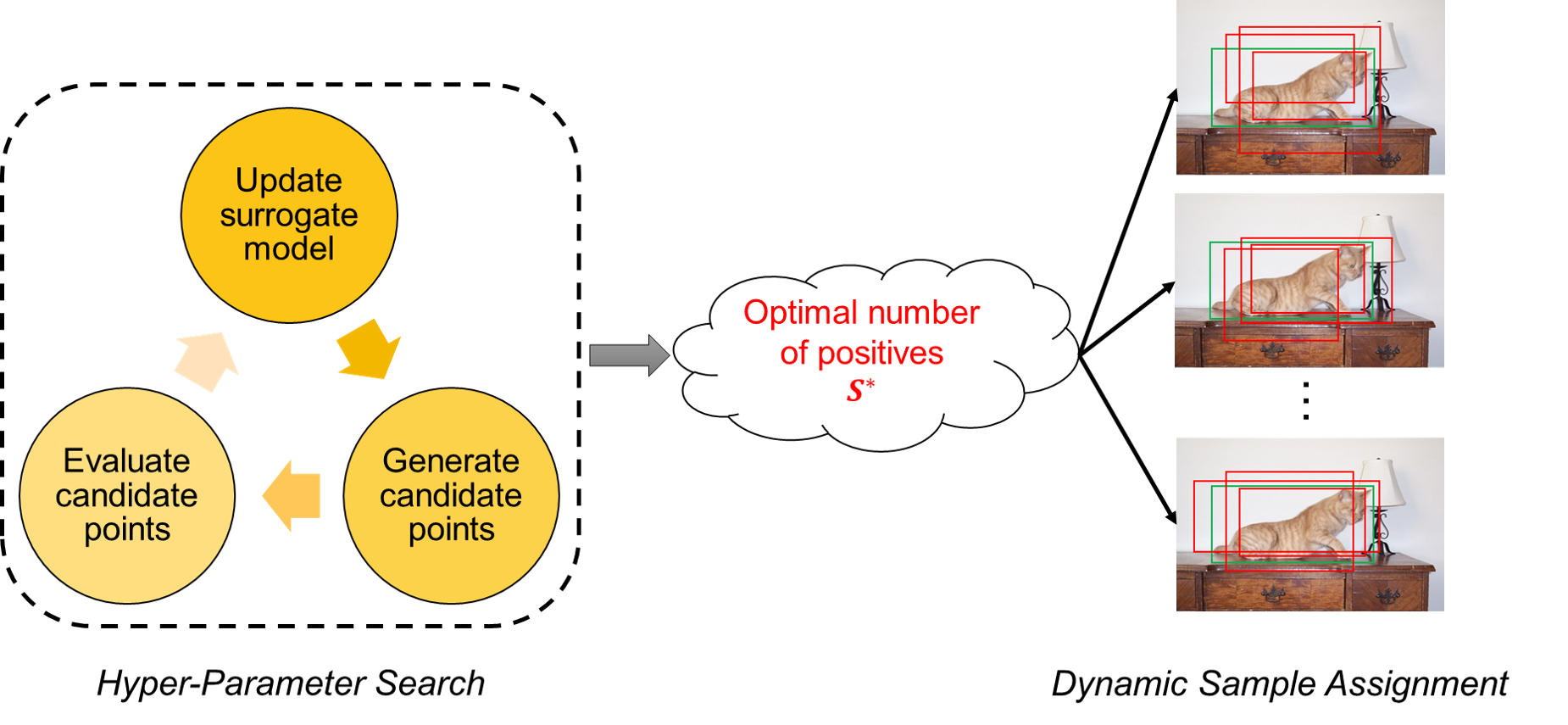}
\end{center}
    \caption{Illustration of our idea of dynamic sample assignment based on hyper-parameter search. First, we define the number of positive samples assigned to each GT as hyper-parameters. Second, we employ a hyper-parameter search algorithm based on surrogate optimization to derive the optimal number of positive samples, including the iterative steps of updating the surrogate model, generating and evaluating candidate points. Then, we propose a dynamic sample assignment procedure to dynamically select the optimal number of positive samples during training.}
\label{fig:overview}
\end{figure}

\section{Introduction}
\label{sec:intro}

Object detection is a fundamental task in the field of computer vision which predicts the objects with locations and pre-defined categories in a given image. It has far reaching impact on wide practical applications include intelligent surveillance, autonomous driving, etc. With the development of deep neural networks, the performance of object detection has been improved by different architectural designs \cite{simonyan2014very,Szegedy_2016_CVPR,he2016resnet} and detection pipelines \cite{girshick2015fast,ren2015faster,liu2016ssd,redmon2018yolov3,tian2019fcos,zhou2019objects,carion2020end}. Two-stage detectors \cite{girshick2015fast,ren2015faster,li2017light} first generate a limited number of candidate proposals for foreground objects (e.g., by region proposal network (RPN)), and then pass these proposals to the network to refine object categories and locations. Two-stage detectors have shown impressive performance but often require large computation overhead. To improve computational efficiency, one-stage methods \cite{liu2016ssd,redmon2016you} have been developed to directly predict object categories and locations in given images without the refinement step. One-stage methods generally can be divided into
anchor-based and anchor-free detectors. Anchor-based detectors~\cite{liu2016ssd,redmon2016you} define dense anchors that are tiled across the image, and directly predict object categories and refine the coordinates of anchors. Anchor-free methods \cite{tian2019fcos,zhou2019objects,kong2020foveabox} eliminate the requirements of pre-defined anchor boxes and typically use center or corner points to define positive proposals and predict offsets to obtain final detections.



Sample assignment plays a prominent part in these object detectors. For anchor-based methods, SSD \cite{liu2016ssd} and RetinaNet \cite{lin2017focal} adopt IoU threshold criterion to divide samples into positives and negatives. YOLOv2 \cite{redmon2017yolo9000} and YOLOv3 \cite{redmon2018yolov3} treat the anchors having the maximum IoU with ground truth (GT) as positive samples. For anchor-free methods, FoveaBox \cite{kong2020foveabox} and FCOS \cite{tian2019fcos} adopt points in the center or entire region of GT bounding box as positive samples. 
We assume that the manually designed assignment strategies have two main limitations. (1) They do not establish the relationships between sample assignment and final detection performance. (2) Positive samples remain unchanged once selected throughout the training process owing to the fixed criteria.


To alleviate these issues, we propose a novel dynamic sample assignment scheme based on hyper-parameter search. Specifically, we cast determining the number of positive samples assigned to each ground truth as a hyper-parameter search problem. We directly define the optimization objective based on the final detection performance (i.e., average precision (AP)). As there exist infinite hyper-parameter choices and the computation cost of training an object detector with a specific hyper-parameter is expensive, we employ a surrogate optimization algorithm (e.g., based on radial basis functions (RBF)) to derive an optimal solution, i.e., the optimal number of positive samples assigned to each GT. We then present a dynamic sample assignment procedure to dynamically select the optimal number of positive samples at each training iteration. Such design brings two advantages: 
(1) We explicitly build the relationship between sample assignment and object detection performance. Through the black-box optimization in hyper-parameter search, the AP value can give feedback to the sample assignment strategy for iterative refinement. (2) We dynamically select positive samples at each training iteration based on a combination of classification and regression losses, rather than fixing positive selection throughout the training process. Figure \ref{fig:overview} illustrates our idea of dynamic sample assignment based on hyper-parameter search.

Our main contributions can be summarized as follows. 
\begin{itemize}
\item We cast determining the number of positive samples assigned to each ground truth as a hyper-parameter search problem, and employ a surrogate optimization algorithm to derive the optimal hyper-parameters efficiently. 

\item Based on our hyper-parameter definition, we design a dynamic sample assignment procedure to dynamically select positives at each training iteration. 


\item We demonstrate the superiority of the proposed method over prior manual sample assignment strategies by applying HPS-Det into different object detection baselines. Particularly, HPS-Det improves RetinaNet-R50 by 2.4\% AP and improves FCOS-R50 by 1.5\% AP on the COCO 2017 validation set, without introducing extra computation overhead during inference.


\item We analyze the hyper-parameter resuability when transferring between different datasets and between different backbones. Improved performance can still be obtained by using proxy dataset or backbone for searching, which can eliminate the demand of repeated search processes. 
\end{itemize}

\section{Related Work}
\subsection{Generic Object Detection}
Generic object detection has achieved impressive performance owing to the advancement of deep neural networks. The early emerging two-stage detectors \cite{ren2015faster,li2017light} first generate candidate region proposals and then refine them to obtain final predictions. While the two-stage
methods have shown promising results, they often consume heavy computation load that limits the applications on resource-constrained
devices in practice. One-stage methods \cite{liu2016ssd,redmon2018yolov3,tian2019fcos,zhou2019objects} thus have been developed for efficient detection without region proposal generation, which can be divided into anchor-based and anchor-free detectors. Anchor-based detectors~\cite{liu2016ssd,redmon2016you,redmon2018yolov3} define dense anchors that are tiled across the image, and directly predict object categories and refine the coordinates of anchors. Anchor-free methods \cite{tian2019fcos,zhou2019objects,kong2020foveabox} eliminate the requirements of pre-defined anchor boxes and typically use center or corner points to define positive proposals and predict offsets to obtain final detections. Recent Transformer-based object detection methods \cite{carion2020end,zhu2021deformable} have been proposed based on the self-attention mechanisms and have shown outstanding performance. Our work is built upon the mainstream CNN-based object detection frameworks (e.g., RetinaNet, FCOS, ATSS).

\subsection{Sample Assignment} 
Anchor-based methods place different sizes of dense anchors on each location of the output feature maps and regress the offset of the bounding box related to these anchors. SSD \cite{liu2016ssd} and RetinaNet \cite{lin2017focal} are anchor-based detectors and adopt similar sample assignment during the training process. Specifically, Intersection-over-Union (IoU) between an anchor and a ground-truth bounding box is utilized as a metric. An anchor is assigned as positive if its IoU with the GT box exceeds a certain threshold, otherwise it is assigned as negative. 
Anchor-free methods predict the keypoints / center and regress the bounding box directly. FCOS \cite{tian2019fcos} takes all points located inside the GT box as positive samples, regresses distances to the four bounding box boundaries and adds a centerness branch for emphasizing points near the center of objects. FoveaBox \cite{kong2020foveabox} defines the positive samples located inside a shrinking area of the GT box shrinking with a factor and negative samples located outside an area with another factor, and other undefined area is ignored during training. ATSS \cite{zhang2019bridging} reveals that the performance gap between anchor-based and anchor-free methods is induced by various definitions of positive and negative samples. It leverages the mean and standard deviation of IoU values between candidate positives and each ground truth to set the selection threshold. For each GT, the threshold may be different but remains fixed during training.
Recent methods explore more strategies by maximum likelihood estimation \cite{zhang2019freeanchor}, probabilistic anchor assignment \cite{kim2020probabilistic}, differentiable label assignment \cite{zhu2020autoassign} and optimal transport assignment \cite{ge2021ota}.
Different from prior methods, we propose to cast determining the number of positives for each GT as a hyper-parameter search problem and dynamically select positive samples during training.


\subsection{Hyper-Parameter Search}
Hyper-parameter search is usually adopted to solve a black-box optimization problem where derivatives of the objective functions are not available. Popular methods include grid search, random search \cite{bergstra2012random} and surrogate optimization \cite{snoek2012practical,regis2013combining,muller2014influence}. Grid search is computationally expensive and only suitable for small amount of hyper-parameters. Random search is more efficient than trials on a grid but may produce unstable results due to randomness. Surrogate optimization can approximate the objective function and
help accelerate convergence to an optimal solution in an efficient way. Typical choices of surrogate models are based on radial basis functions (RBF) \cite{regis2013combining},  Gaussian process (GP) \cite{snoek2012practical}, etc. Recently, AABO \cite{ma2020aabo} proposes an adaptive anchor box optimization method for object detection via Bayesian sub-sampling, which can enhance the performance of anchor-based detection baseline. Our method differs from AABO in two aspects. (1) AABO aims to search for the anchor configurations, while we aim to search for the number of positive samples. (2) AABO is only applied to anchor-based methods, while our method is compatible to both anchor-based and anchor-free detectors.

\section{Proposed Approach}

\subsection{Hyper-Parameter Definition} 
Suppose we have $N$ ground-truth bounding boxes in total in the training set, the number of positive samples assigned to each GT $g$ can be denoted as $p(g_i), i=1,2,\dots,N$, where $p(\cdot)$ is often manually designed in most of existing detectors. For instance, in RetinaNet, $p(\cdot)$ is determined by counting how many candidate anchors exceeds an IoU threshold with each GT; In FCOS, $p(\cdot)$ is determined by counting how many candidate points locate inside the GT box; In ATSS, $p(\cdot)$ is determined via an adaptive threshold that relies on the mean and standard deviation of IoU values between candidates and GTs. Our main idea is treating the number of positives assigned to each GT as a hyper-parameter and automatically searching for the optimal choices instead of manual determinations. 
However, if we simply allocate a distinct hyper-parameter for each GT, it will lead to a large amount of hyper-parameters so that the optimization difficulty and computation time of hyper-parameter search will increase. 


Thus, we group ground truths based on their scales (i.e., sizes and aspect ratios) into multiple bins and only define a hyper-parameter for each bin. The hypothesis is that we allocate the same number of positives for ground truths with similar scales. Specifically, to handle the size variation, we match each GT to a best pyramid level in FPN and set different hyper-parameters for different levels. To handle the aspect ratio variation, we simply divide several ranges and set different hyper-parameters for different ranges. 
More formally, the complete hyper-parameter list to be searched can be defined as:
\begin{equation}
    S=[s^1_1,\cdots,s_{k_1}^1,\cdots,\underbrace{s^i_1,\cdots,s_{k_i}^i}_{i\text{-th level}},\cdots,s^L_1,\cdots,s_{k_L}^L]
\label{eq:hyper}
\end{equation}
Here, $L$ means the number of pyramid levels. $s^i_j$ means the number of positive samples when a ground truth $g$ matches the $i$-th pyramid level and its aspect ratio falls into the $j$-th range.
In this way, our hyper-parameter definition method can largely reduce the amount of hyper-parameters and promote the search process.

\subsection{Surrogate Optimization}

Based on the above hyper-parameter definition, we employ surrogate optimization to seek the optimal hyper-parameters. The brute-force search process is computationally expensive as (1) the search space of hyper-parameters is huge and (2) evaluating if a certain hyper-parameter is good or not requires training the detector, which costs high computation overhead. Surrogate optimization is a kind of efficient black-box optimization algorithm for hyper-parameter search.
The core idea is to build a low-cost surrogate model to approximate the objective function and solve an auxiliary problem to select promising hyper-parameter candidates for next evaluations. The procedure of evaluating the candidate points, updating the surrogate model and generating the new points is repeated until a stopping criterion has been met. 

Specifically, the main procedure of our surrogate optimization algorithm for object detection has the following steps.
\begin{enumerate}
    \item Initialize a small dataset $B$ of pair data $\{(S_i, \mathcal{F}(S_i)\}$, where $S_i$ is the hyper-parameter instance and $\mathcal{F}(\cdot)$ is the evaluation function for object detection (i.e., AP).
    \item Build a surrogate model $\hat{\mathcal{M}}$ based on RBF to fit $B$: $\hat{\mathcal{M}}(S_i) \to \mathcal{F}(S_i)$.
    \item Determine the next promising candidate hyper-parameters (i.e., points in the high-dimensional search space) based on the surrogate model $\hat{\mathcal{M}}$.
    \item Evaluate the new candidate hyper-parameter using $\mathcal{F}(\cdot)$ and update $B$ with the new pair data.
    \item Repeat the above three steps until the maximum number of iterations is met and return the optimal hyper-parameter instance with best evaluation result.
\end{enumerate}

\begin{algorithm}[t!]
\raggedright
\caption{Surrogate Optimization on Detector}
\label{alg:algorithm}
\textbf{Input}: \\
$\mathcal{M}$ is the object detector \\
$D_\text{train}$ is a sampled subset of training dataset \\ 
$D_\text{val}$ is the validation dataset \\ 
$\mathcal{F}$ is the evaluation function (i.e., detection AP) \\
$\mathbb{S}$ is the search space of defined hyper-parameter in Eq. \ref{eq:hyper} \\
\textbf{Output}: \\
$S^*$ is the optimal hyper-parameter instance \\
\begin{algorithmic}[1] 
\STATE Intialize a small dataset of pair data by evaluating the detector: $B=\{(S_i, \mathcal{F}(\mathcal{M}(D_\text{train}), D_\text{val}, S_i))\}$, $S_i \in \mathbb{S}$
\STATE Build a surrogate model $\hat{\mathcal{M}}$ to fit $B$\\
\STATE Let $t=0$
\WHILE{$t<$ max\_iter}
\STATE {$S \gets $ select the next promising candidate hyper-parameter based on $\hat{\mathcal{M}}$} \\
\STATE {$B = B \cup (S, \mathcal{F}(\mathcal{M}(D_\text{train}), D_\text{val}, S)$) } \\
\STATE {Update $\hat{\mathcal{M}}$ with $B$} \\
\STATE {$t \gets t+1$} \\
\ENDWHILE
\STATE {$S^* \gets$ get the hyper-parameter instance with the highest AP value in $B$}    \\
\STATE \textbf{return} $S^*$
\end{algorithmic}
\end{algorithm}

Algorithm \ref{alg:algorithm} describes our surrogate optimization algorithm for object detection in detail. To further speed up the hyper-parameter optimization, we randomly sample a small subset of training data for the search procedure and empirically find it works well. 

\subsection{Dynamic Sample Assignment}
After obtaining the optimal hyper-parameters, we design a dynamic sample assignment (DSA) scheme to dynamically select the optimal number of positive samples assigned to each GT during training. RetinaNet \cite{lin2017focal} and FCOS \cite{tian2019fcos} are representative anchor-based and anchor-free detectors respectively, both of which rely on manual sample assignment. We take these two baseline detectors as examples to introduce our DSA algorithm: 

\begin{algorithm}[t!]
\raggedright
\caption{Dynamic Sample Assignment}
\label{alg:sample definition algorithm}
\textbf{Input}: \\
$G$ is a set of ground-truth boxes of an image \\
$L$ is the number of pyramid levels \\
$A$ is a set of all anchor boxes / points and $A_i$ is the subset from $i$-th pyramid level \\
$S^*$ is the searched hyper-parameter list \\
\textbf{Output}: \\
$P$ is a set of positive samples\\
$N$ is a set of negative samples \\
\begin{algorithmic}[1] 
\STATE {$P \gets \emptyset $}
\FOR{each ground truth $g \in G$}
\STATE {$C_{g} \gets \emptyset$}\\
\FOR{each level $i \in [1, L]$}
\STATE{$C_i \gets $ select anchors from A$_i$ with IoU scores $>0.1$ for RetinaNet, or select points from A$_i$ in the entire region of $g$ if this level meets the scale constraint else $C_i=\emptyset$ for FCOS}
\STATE {$C_{g} = C_{g} \cup C_{i} $}
\ENDFOR
\STATE {Compute classification loss: $l_{cls} = \text{FocalLoss}(C_{g}, g)$ \\
}
\STATE {Compute regression loss: $l_{reg} = \text{IoULoss}(C_{g}, g)$ \\
}
\STATE {Compute combined loss: $l = l_{cls} + \alpha l_{reg}$ \\
}
\STATE {$i^* \gets$ determine the best matched pyramid level of $g$} \\
\STATE {$j^* \gets$ determine the aspect ratio bin of $g$} \\
\STATE {$s_{j^*}^{i*} \gets$ find the optimal number of positives assigned to $g$ by index in $S^*$} \\
\STATE {$P_g \gets $ GetTopKSampleWithLoss($C_g, l, s_{j^*}^{i*})$} \\
\STATE {$P=P \cup P_{g}$} \\
\ENDFOR
\STATE {$N \gets A - P$} \\
\STATE \textbf{return} $P, N$
\end{algorithmic}
\end{algorithm}

\begin{enumerate}
    \item For each ground truth $g$, build an initial set of candidate positive samples $C_g$. For RetinaNet, we select anchors that have higher IoU scores than a small threshold. For FCOS, we select points in the entire region of ground truth. 
    \item Compute the classification and bounding box regression losses of candidate positive samples.
    \item Determine the best matched pyramid level $i^*$ of $g$. For RetinaNet, we determine $i^*$ where an anchor has the maximum IoU score with $g$. For FCOS, we determine $i^*$ where the bounding box regression offsets of a point meet the scale constraint as in \cite{tian2019fcos}.
    \item Compute the aspect ratio of $g$ and determine the bin $j^*$ it falling into. 
    \item Select $s_{j^*}^{i^*}$ positive samples from $C_g$ with the lowest combined losses. 
\end{enumerate}

Algorithm \ref{alg:sample definition algorithm} describes our dynamic sample assignment procedure in detail. The proposed algorithm brings three main merits for object detection. First, unlike manual or fixed sample assignment, we dynamically select positive samples according to the searched hyper-parameters, which can facilitate the optimization of the detection network. Second, our method keeps the same inference pipeline as the original detectors and thereby will not introduce extra computation overhead during inference. Third, the proposed dynamic sample assignment scheme is compatible to the mainstream anchor-based and anchor-free detectors.

\section{Experiments}
\subsection{Experimental Settings}  

\textbf{Datasets and Metrics.} We conduct extensive experiments to validate the effectiveness of our proposed method on the MS COCO 2017 and VOC 2007 datasets. MS COCO 2017 has 80 object classes, containing 118k training, 5k validation and 21k testing images. VOC 2007 has 20 object classes, containing 2.5k training, 2.5k validation and 5k testing images. For MS COCO 2017, we train detectors on the train split and evaluate the performance on the val and test-dev splits. For VOC 2007, we train detectors on the train+val split and evaluate the performance on the test split. We report the performance with standard MS COCO AP and VOC MAP@0.5.

\begin{table*}[t!] 
  \begin{center}
  \begin{tabular}{l | c | c | c | c | c }
    \toprule
     Baselines & FPN level-1 &  FPN level-2  &   FPN level-3  &  FPN level-4  &  FPN level-5 \\
    \midrule
    RetinaNet  & 11,11,20 & 44,58,51 & 40,70,60 & 41,55,80 & 47,77,73  \\ 
    FCOS  & 17,15,15  &  20,15,25 & 15,17,21 & 16,25,19 & 20,4,3 \\ 
    ATSS & 7,11,12 & 27,29,25  & 20,20,22   & 18,22,31  &3,31,24  \\

    \bottomrule
  \end{tabular}
  \end{center}
    \caption{Searched 15-dim hyper-parameter instances for different detectors using the ResNet-50 backbone on COCO.
    }
      \label{table:optimalhp}
\end{table*}

\begin{figure}[t!]
\footnotesize
\begin{center}
\begin{tabular}{@{}cc@{}}
\includegraphics[width = 0.48\linewidth]{{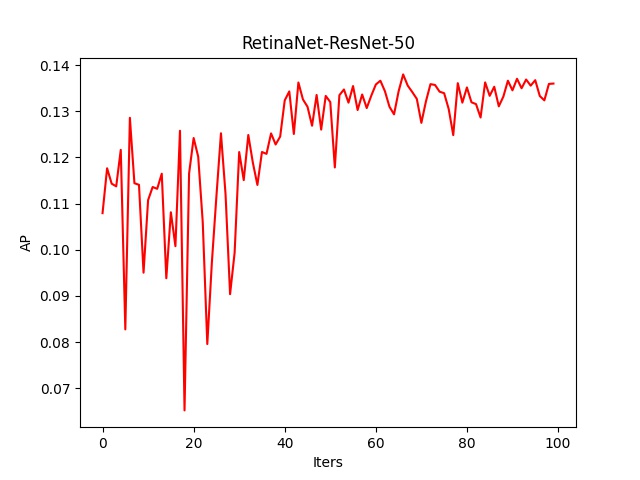}} & 
\includegraphics[width = 0.48\linewidth]{{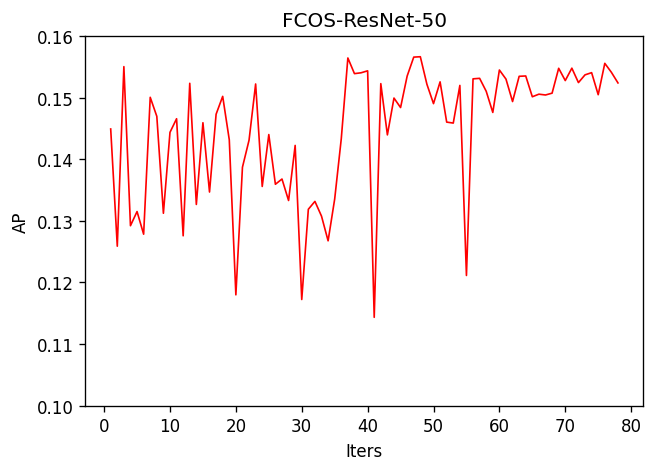}} \\
{\footnotesize (a) RetinaNet} & 
{\footnotesize (b) FCOS}
\end{tabular}
\end{center}
\caption{AP curves during the search process based on the RetinaNet and FCOS baselines with ResNet-50.}
\label{figure:ap curves}
\end{figure}

\textbf{Detection Baselines.} We take three mainstream object detectors as our baselines: RetinaNet, FCOS and ATSS. We use their official code for re-implementation, all of which are based on maskrcnn-benchmark. Unless specified otherwise, we report the results of vanillar FCOS without additional tricks. We use the pre-trained networks on ImageNet-1K with FPN as backbones for detection.

\textbf{Training Details.} By default, images are resized to a maximum scale of 800$\times$1333 without changing the aspect ratio. The whole detector is trained with the Stochastic Gradient Descent (SGD) with momentum of 0.9, weight decay of 0.0001 and batch size of 16. We train the network with 12 epochs in total for single-scale training. The learning rate is initialized as 0.01 and decayed by 10$\times$ at 8-th and 11-th epoch. Our experiments are conducted on PyTorch with 4 V100/P100 GPUs.

\textbf{Inference Details.} We adopt the standard single-scale inference setting for evaluations.
The network outputs the prediction of bounding boxes and their corresponding class probabilities.
We then filter out a large number of background bounding boxes by a confidence score threshold of 0.05 and keep top 1k candidate boxes per feature pyramid level. Non-Maximum Suppression (NMS) with IoU threshold of 0.5 is applied to yield the final detection results.

\subsection{Results of Hyper-Parameter Search}
\textbf{Hyper-Parameter Configurations.} The backbone used in each detector has 5 FPN levels. By default, we define 3 hyper-parameters per level by dividing the aspect ratio into 3 bins: $(-\infty,1/2]$, $(1/2,2]$, $(2,+\infty)$. It finally results in a 15-dim hyper-parameter list.


\textbf{Search Process.}
We use the surrogate optimization framework pySOT \cite{eriksson2019pysot} and DYCORS strategy \cite{regis2013combining} for hyper-parameter search. 
To reduce the search space, the maximum of each hyper-parameter is set to 80 for RetinaNet, 25 for FCOS and 32 for ATSS. To accelerate the search process, we randomly sample 5k images of COCO and perform maximum 100 iterations for search.
Figure \ref{figure:ap curves} presents two examples of AP curves during the search process. We observe that AP increases and converges within a small amount of iterations. We hypothesize that the fluctuating phenomenon in early iterations may be caused by the hyper-parameter sensitivity as the number of positives samples assigned to GTs has a large influence on the detection performance.

\textbf{Searched Results.} Our searched hyper-parameters using the ResNet-50 backbone on the subset of COCO are presented in Table \ref{table:optimalhp}. Each iteration of the search step costs $\sim$1 hour in our experiments. The hyper-parameters searched by ResNet-50 can be directly applied to other larger backbones, which can reduce the computation overhead of repeated search step. 

\begin{table}[!t] \small
  \begin{center}
  \begin{tabular}{l | l | l }
    \toprule
     Methods &   Backbone     & AP (\%) \\
    \midrule
    RetinaNet & ResNet-50 & 35.8   \\
    RetinaNet + HPS-Det & ResNet-50 & 38.2 (\textcolor{red}{+2.4}) \\
    RetinaNet & ResNet-101      & 38.6  \\
    RetinaNet + HPS-Det & ResNet-101 & 40.4 (\textcolor{red}{+1.8}) \\
    RetinaNet & ResNeXt-32$\times$8d-101    & 40.6 \\
    RetinaNet + HPS-Det & ResNeXt-32$\times$8d-101    & 42.7 (\textcolor{red}{+2.1}) \\
    \midrule
    FCOS & ResNet-50 & 36.9 \\
    FCOS + HPS-Det & ResNet-50 & 38.4 (\textcolor{red}{+1.5})  \\
    FCOS & ResNet-101 & 39.1  \\
    FCOS + HPS-Det & ResNet-101 & 40.4 (\textcolor{red}{+1.3}) \\
    FCOS & ResNeXt-32$\times$8d-101& 41.1  \\
    FCOS + HPS-Det & ResNeXt-32$\times$8d-101 & 42.5 (\textcolor{red}{+1.4})  \\
    \bottomrule
  \end{tabular}
  \end{center}
    \caption{Detection performance comparisons (\%) on the COCO 2017 validation set with different baselines. 
    All the results are obtained in the same single-scale training with 1$\times$ scheduler and single-scale inference settings.
    }
    \label{table:basemodel}
\end{table}

\begin{table}[!t] 
  \begin{center}
  \begin{tabular}{l | l | l }
    \toprule
     Methods &   Backbone     & AP (\%) \\
    \midrule
    RetinaNet & Swin-T & 37.8   \\
    RetinaNet + HPS-Det & Swin-T & 39.9 (\textcolor{red}{+2.1}) \\
    RetinaNet & Swin-S      & 39.6  \\
    RetinaNet + HPS-Det & Swin-S & 41.7 (\textcolor{red}{+2.1})\\
    RetinaNet & Swin-B    & 40.1 \\
    RetinaNet + HPS-Det & Swin-B    & 41.9 (\textcolor{red}{+1.8})\\
    RetinaNet & Swin-B$^*$    & 40.6 \\
    RetinaNet + HPS-Det & Swin-B$^*$    & 42.8 (\textcolor{red}{+2.2})\\
    \midrule
    FCOS & Swin-T & 39.6 \\
    FCOS + HPS-Det & Swin-T & 40.5 (\textcolor{red}{+0.9}) \\
    FCOS & Swin-S & 41.2  \\
    FCOS + HPS-Det & Swin-T & 42.2 (\textcolor{red}{+1.0}) \\
    FCOS & Swin-B & 41.3  \\
    FCOS + HPS-Det & Swin-B & 42.4 (\textcolor{red}{+1.1}) \\
    FCOS & Swin-B$^*$ & 41.2  \\
    FCOS + HPS-Det & Swin-B$^*$ & 42.8 (\textcolor{red}{+1.6}) \\
    \bottomrule
  \end{tabular}
  \end{center}
    \caption{Detection performance comparisons (\%) on the COCO 2017 validation set with different Swin Transformer backbones. 
    We re-implement the improved FCOS baseline in this experiment. We use 1$\times$ scheduler for Swin-T and Swin-S, and use 2$\times$ scheduler for Swin-B. $^*$ indicates multi-scale training.
    }
    \label{table:swin}
\end{table}

\begin{table*}[t!] 
  \begin{center}
  \begin{tabular}{l | c | c | c|c|c|c|c}
    \toprule
     Methods &   Backbone     & AP & AP$_{50}$ & AP$_{75}$ & AP$_S$ & AP$_M$ & AP$_L$ \\
    \midrule
    \textit{Anchor-Based Detectors:} & & & & & & & \\
    SSD513 \cite{liu2016ssd} & ResNet-101 & 31.2 & 50.4 & 33.3 & 10.2 & 34.5 & 49.8 \\
    YOLOv3 (608$\times$608) \cite{redmon2018yolov3} & Darknet-53 & 33.0 & 57.9 & 34.4 & 18.3 & 35.4 & 41.9 \\
    Faster R-CNN w/ FPN \cite{lin2017feature} & ResNet-101 & 36.2 & 59.1 & 39.0 & 18.2 & 39.0 & 48.2 \\
    PISA \cite{cao2019prime} & ResNet-50 & 37.3 & 56.5 & 40.3 & 20.3 & 40.4 & 47.2 \\
    Mask R-CNN \cite{he2017maskrcnn} & ResNet-101 & 38.2 & 60.3 & 41.7 & 20.1 & 41.1 & 50.2 \\
    RetinaNet \cite{lin2017focal} & ResNet-50 & 35.7 & 55.0 & 38.5 & 18.9 & 38.9 & 46.3 \\
    RetinaNet \cite{lin2017focal} & ResNet-101 & 37.8 & 57.5 & 40.8 & 20.2 & 41.1 & 49.2 \\
    Noisy Anchors \cite{li2020learning} & ResNet-101 & 41.8 & 61.1& 44.9 & 23.4 &44.9 & 52.9 \\
    FreeAnchor* \cite{zhang2019freeanchor} & ResNeXt-64$\times$4d-101 & 44.9 & 64.3& 48.5 & 26.8 &48.3 & 55.9 \\
    \midrule
    \textit{Anchor-Free Detectors:} & & & & & & & \\
    FoveaBox \cite{kong2020foveabox} &ResNet-50 &37.1 &57.2 & 39.5&21.6 &41.4 &49.1 \\
    FCOS \cite{tian2019fcos} & ResNet-50&37.1&55.9 &39.8 &21.3 &41.0 &47.8 \\
    FSAF \cite{zhu2019feature} &ResNet-50 &37.2 & 57.2&39.4 &21.0 & 41.2&49.7 \\
    ExtremeNet \cite{zhou2019extremenet} & Hourglass-104 &40.2 &55.5 & 43.2&20.4 &43.2 &53.1 \\    
    CornerNet \cite{law2018cornernet} & Hourglass-104 &40.5 &56.5 & 43.1&19.4 &42.7 &53.9 \\    
    CenterNet-HG \cite{zhou2019objects} & Hourglass-104 &42.1 &61.1 & 45.9&24.1 &45.5 &52.8 \\    
    ATSS \cite{zhang2019bridging} & ResNet-50 & 39.3 & - & - & - & - & - \\
    \midrule
    \textit{Ours:} & & & & & & & \\
    RetinaNet + HPS-Det & ResNet-50 &   38.6  & 56.3 & 41.6 & 19.3 & 42.0 & 50.9     \\
    RetinaNet + HPS-Det & ResNet-101 &  40.8 &58.6  &44.0  &20.6 &44.6 &53.6 \\
    RetinaNet + HPS-Det & ResNeXt-32$\times$8d-101 &   43.1 & 61.5 &46.6  &23.3 &46.8 &56.0  \\
    FCOS + HPS-Det & ResNet-50 & 38.5 &56.2 & 41.6 & 20.3 &41.5 &49.5    \\
    FCOS + HPS-Det & ResNet-101 &  40.5 &58.5  &43.9  &21.6 &43.6 & 52.0 \\
    FCOS + HPS-Det & ResNeXt-32$\times$8d-101  &42.8   & 61.3 & 46.3 &24.4 & 45.9&53.8 \\
    ATSS + HPS-Det & ResNet-50 &   39.9  & 57.1  &43.3 &20.1 &43.7 & 52.1  \\       
    ATSS + HPS-Det & ResNet-101 &  42.3  & 59.5 &46.0  &21.8 &46.4 &  54.5 \\   
    ATSS + HPS-Det & ResNeXt-64$\times$4d-101 &   45.3 & 63.5 &49.3  &25.2 & 49.7& 57.8  \\ 
    \bottomrule
  \end{tabular}
  \end{center}
    \caption{Detection performance comparisons (\%) on the COCO 2017 test-dev set. 
    For fair comparisons, all the results are obtained in the same single-scale inference settings. * means 2$\times$ multi-scale training.
    }
    \label{table:sota}
\end{table*}

\begin{table*}[t!] \footnotesize
  \begin{center}
  \begin{tabular}{ c | c | c | c c c }
    \toprule
      \multirow{2}{*}{Proxy Backbone} & \multirow{2}{*}{Hyper-Param Config} & \multirow{2}{*}{Searched Results} & \multicolumn{3}{c}{Target Backbone} \\
      & & & ResNet-50 & ResNet-101 & ResNeXt-32$\times$8d-101 \\
    \midrule
     \xmark & \xmark & \xmark & 36.9 & 39.1 & 41.1 \\
     ResNet-50 & (3,3,3,3,3) & 17,15,15,20,15,25,15,17,21,16,25,19,20,4,3 & 38.4 & 40.4 & 42.5 \\
     ResNet-50 & (1,1,1,1,1) & 5,19,25,24, 20 & 37.7 &	39.9 &	42.5 \\
     ResNet-50 & (5) & 10,16,16,8,25 & 38.6 & 40.5 & 42.8 \\
     ResNet-50 & (3) & 16,25,21 & 38.7 & 41.2 & 43.1 \\
     ResNet-50 & (1) & 24 & 38.4 & 40.8 & 43.1 \\
    
    \bottomrule
  \end{tabular}
  \end{center}
    \caption{Ablation studies on hyper-parameter configurations. We use FCOS as the baseline and use ResNet-50 as proxy backbone to search for the hyper-parameters in this experiment.
    }
      \label{table:config}
\end{table*}

\begin{table}[!t] \footnotesize
  \begin{center}
  \begin{tabular}{l | c | c | c }
    \toprule
     Methods & Backbone & Search Dataset & MAP (\%)  \\
    \midrule
    FCOS & ResNet-50 & \xmark &  69.68  \\
    FCOS + HPS-Det & ResNet-50 & VOC-0.5k &  70.45   \\
    FCOS + HPS-Det & ResNet-50 & VOC-1k &   70.92 \\
    FCOS + HPS-Det & ResNet-50 & VOC-2.5k &  71.70  \\
    FCOS + HPS-Det & ResNet-50 & COCO-5k &  70.81    \\
    \bottomrule
  \end{tabular}
  \end{center}
    \caption{Detection performance on the VOC 2007 test set. 
    }
      \label{table:voc}
\end{table}

\begin{table}[t!] \footnotesize
  \begin{center}
  \begin{tabular}{l | c | c | c }
    \toprule
    Proxy Backbone & Target Backbone & AP (\%) & Search Time \\
    \midrule
    ResNet-18 & ResNet-50 & 38.1 & 0.9h \\
    ResNet-18 & ResNet-101 & 40.1 & 0.9h \\
    ResNet-50 & ResNet-50 & 38.4 & 1.2h \\
    ResNet-50 & ResNet-101 & 40.4 & 1.2h\\
    ResNet-101 & ResNet-101 & 40.7 & 1.7h \\
    \bottomrule
  \end{tabular}
  \end{center}
    \caption{Experiments on hyper-parameter reusability between different backbones using FCOS. We compare the AP and search time per iteration on COCO.
    }
      \label{table:r18}
\end{table}

\subsection{Results of Object Detection}

\textbf{Comparisons to Different Detection Baselines.}
Table \ref{table:basemodel} compares our HPS-Det with different detection baselines on the COCO 2017 validation set. We replace their sample assignment strategies with our dynamic sample assignment based on the optimal number of positives derived by hyper-parameter search. Results show that our method can obtain improved performance over different detection frameworks with different network backbones. Particularly, HPS-Det improves the anchor-based RetinaNet by 2.4\% AP and improves the anchor-free FCOS by 1.5\% AP with the same ResNet-50. Even with the larger backbones (e.g., ResNet-101 or ResNeXt-32$\times$8d-101), we can surpass both baselines with a large margin. 
Compared to the strong baseline of ATSS which adopts an adaptive sample selection strategy, HPS-Det obtains slightly better performance (39.2\% vs. 39.7\%) with the same ResNet-50 backbone, which further exhibits the effectiveness of our method.

\textbf{Results on Transformer-based Backbones.}
Table \ref{table:swin} compares our HPS-Det with different Transformer-based backbones on the COCO 2017 validation set. We implement the anchor-based RetinaNet and anchor-free FCOS baselines with the Swin Transformer backbones, which are pre-trained on ImageNet-1K. The results show that our HPS-Det can obtain improved performance over all the Transformer backbones, which further validate the superiority of our method. 


\textbf{Comparisons to the State-of-the-Arts.} 
Table \ref{table:sota} compares our method with the state-of-the-art object detectors on the COCO test-dev set. Although the hyper-parameters are optimized on the validation set, the final detectors show similar improvements over the anchor-based and anchor-free baselines on COCO test-dev. For example, we improve RetinaNet, FCOS, ATSS by 2.9\%, 1.4\%, 0.6\% AP with ResNet-50, respectively. 
By applying HPS-Det on ATSS with the ResNeXt-64$\times$4d-101 backbone, we obtain a single-model single-scale AP of 45.3\%, which is competitive with the state-of-the-art methods. 

\textbf{Comparisons to Prior Sample Assignment Methods.} We have validated the superiority of our method over three different detection baselines that rely on fixed / adaptive sample assignment. We compare more related sample assignment methods for object detection. Compared to FreeAnchor \cite{zhang2019freeanchor} that adopts a learning-to-match approach to break IoU restriction, our AP obtained by 1$\times$ single-scale training is even higher (+0.4\%) than its AP by 2$\times$ multi-scale training with the same ResNeXt-64$\times$4d-101 backbone on COCO test-dev. Compared to the most relevant AABO \cite{ma2020aabo} that adopts Bayesian sub-sampling to search for the anchor configurations, our method achieves AP gains of 0.9\% with the same ReinaNet-ResNet-101 baseline (39.5\% vs. 40.4\%) on COCO validation. Both AABO and our HPS-Det are motivated by hyper-parameter optimization, but the optimization targets are different. AABO aims to search for optimal anchor configurations, while we aim to search for the optimal number of positive samples using the same anchor configurations with the baseline detector. Besides, AABO highly depends on the anchor mechanism and our method can be applied to both anchor-based and anchor-free detection methods. Compared to recent methods \cite{kim2020probabilistic,zhu2020autoassign,ge2021ota} by different label assignment strategies, our HPS-Det can obtain comparable results but with a technically different scheme based on hyper-parameter search for object detection.

\begin{figure*}[!t]
\begin{center}
\includegraphics[width = 0.9\linewidth]{{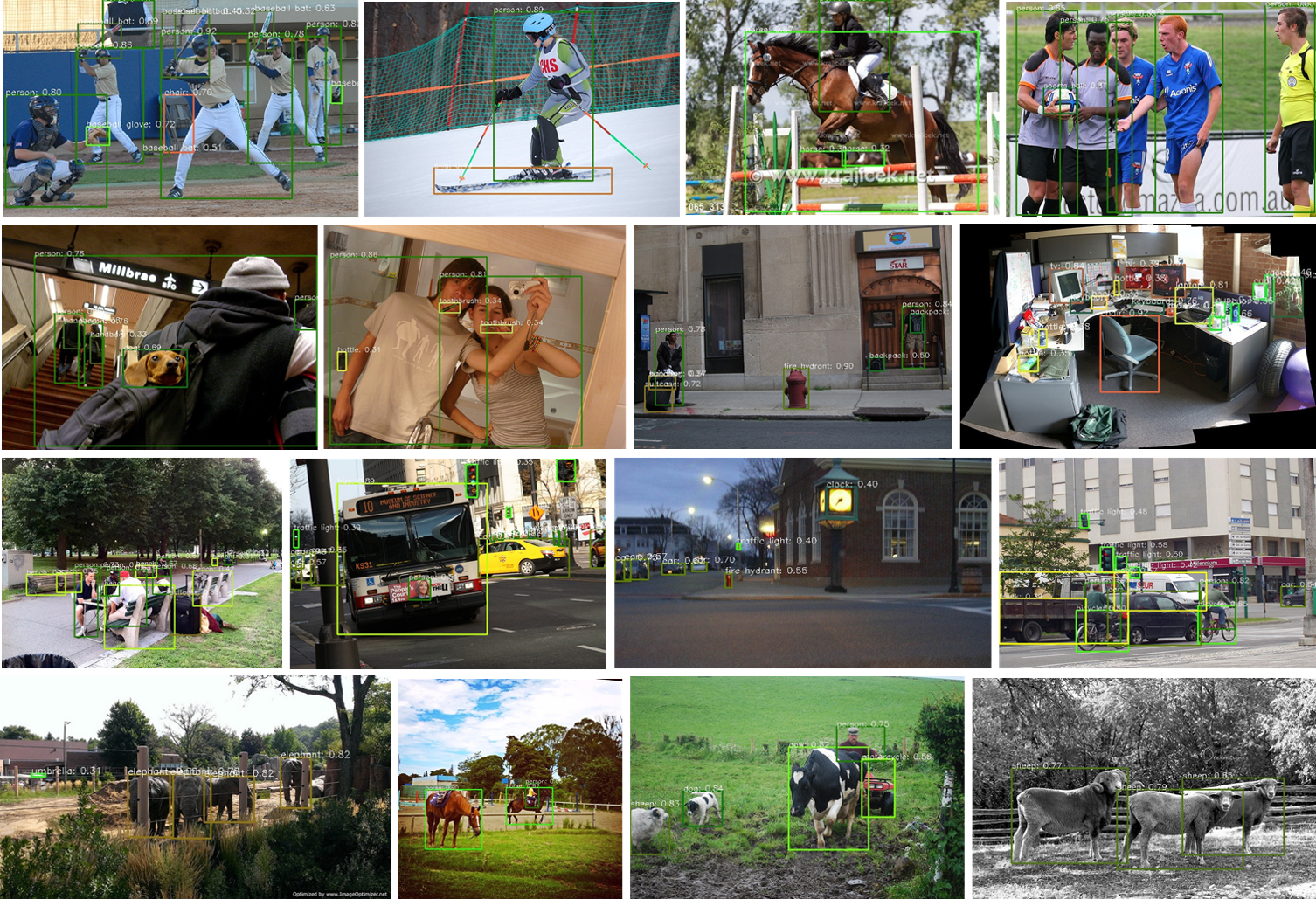}}
\end{center}
\caption{Examples of detection results by our method on COCO.}
\label{figure: detection}
\end{figure*}

\textbf{Qualitative Results}
Figure \ref{figure: detection} shows sampled detection results of our method on COCO. Our HPS-Det can detect objects with varying scales, appreances, etc.

\subsection{Ablation Study}
\textbf{Effect of Hyper-Parameter Configurations.} 
We provide ablation studies to validate the effect of hyper-parameter configurations based on FCOS in Table \ref{table:config}. The config of ``(3,3,3,3,3)'' denotes 5 pyramid levels and 3 hyper-parameters per level (i.e., 3 aspect ratio bins: $(-\infty,1/2], (1/2,2], (2,+\infty)$), which incorporates the priors of GT sizes and aspect ratios to define the 15-dim hyper-parameter list. ``(1,1,1,1,1)'' denotes 5 pyramid levels and 1 hyper-parameter per level, which only incorporates GT sizes to define the 5-dim hyper-parameter list. ``(3)'' defines 3 hyper-parameters by dividing 3 aspect ratio bins, which only considers GT aspect ratios. Also, ``(5)'' indicates the config by dividing more aspect ratio bins (i.e., $(-\infty,1/4], (1/4,1/2], (1/2,2], (2,4], (4,+\infty)$). The extreme case of ``(1)'' means that we allocate the same number of positives for all the GTs, without considering their sizes and aspect ratios. We use ``(3,3,3,3,3)'' as default config for all the baseline detectors in our method. The results show that different configurations outperform the baseline and less hyper-parameters still can achieve promising performance. Taking ResNet-50 as target backbone, we obtain +1.8\% AP gain (36.9\% vs. 38.7\%) over the FCOS baseline with the config of ``(3)''. Particularly, using only one hyper-parameter even performs slightly better than the default config of ``(3,3,3,3,3)'' on ResNet-101 (40.4\% vs. 40.8\%) and ResNeXt (42.5\% vs. 43.1\%), which validates that our hyper-parameter definition does not necessarily rely on the pre-defined GT priors. 

\textbf{Choices of Hyper-Parameter Search Algorithm.}
There exist many hyper-parameter search algorithms, including grid search, random search, genetic algorithm, surrogate model algorithm, etc. Grid search is computationally expensive and random search may produce unstable results due to randomness. Genetic algorithm usually costs more time to converge than surrogate model algorithm. In this work, we adopt radial basis function (RBF) surrogates with dynamic coordinate search (DYCORS) \cite{regis2013combining} for hyper-parameter search. We also conduct ablation experiments to compare different optimization strategies, including Stochastic RBF (SRBF), Expected Improvement (EI), and Lower Confidence Bound (LCB). Taking FCOS-ResNet-50 as baseline, all the strategies can obtain improved detection performance. Baseline: 36.9\%, DYCORS: 38.4\%, SRBF: 38.1\%, EI: 38.6\%, LCB: 38.3\%.


\textbf{Effect of Search Dataset.} 
We evaluate our method on the VOC 2007 dataset using FCOS-ResNet-50 as baseline in Table \ref{table:voc}. With different sizes of search dataset, all of our methods can obtain improved performance over the baseline. The results also show that using more training data for hyper-parameter search can further boost the final detection performance. 

\subsection{Analysis of Hyper-Parameter Reusability}
A limitation of our method is requiring extra search step for object detection. We have made efforts to reduce the training cost by reusing the hyper-parameters between different datasets and between different backbones.  
First, Table \ref{table:voc} shows that applying the hyper-parameters optimized from COCO to VOC still can improve the detection performance (69.68\% $\to$ 70.81\%). Second, Table \ref{table:r18} shows that applying the hyper-parameters optimized from a light-weight backbone to a larger one can still bring performance gains without repeated search process. For example, training FCOS-ResNet-50 with hyper-parameters searched on ResNet-18 performs slightly worse than searching on the target backbone (38.1\% vs. 38.4\%). However, searching on ResNet-18 can save 0.3h per iteration compared to searching on ResNet-50. Table \ref{table:config} also shows that directly using the searched hyper-parameters by ResNet-50 can improve the performance on the larger backbones of ResNet-101 and ResNeXt under different hyper-parameter configurations. These results exhibit the hyper-parameter reusability of our method between different datasets and between different backbones.

\section{Conclusion}
In this work, we propose to cast dertermining the optimal number of positives assigned to GTs as a hyper-parameter optimization problem. We then design a dynamic sample assignment procedure based on the searched hyper-parameters. 
We demonstrate the superiority of our method over prior manual sample assignment strategies in different detection baselines.
We also analyze the hyper-parameter reusability when transferring between different datasets and between different backbones. 
We consider applying our method for more object detection frameworks (e.g., two-stage detectors) as future work. We believe that our idea of introducing hyper-parameter search into object detection is interesting and can inspire new insights to the community.

{\small
\bibliographystyle{ieee_fullname}
\bibliography{egbib}
}










\end{document}